\documentclass{bmvc2k}

\usepackage{bbm}
\usepackage{amssymb}
\usepackage{booktabs}
\graphicspath{{images/}}

\newcommand{\squeezeup}{\vspace{-2.5mm}}

\title{Context Matters: Refining Object Detection in Video with Recurrent Neural Networks}

\addauthor{Subarna Tripathi}{stripathi@ucsd.edu}{1}
\addauthor{Zachary C. Lipton}{zlipton@cs.ucsd.edu}{1}
\addauthorr{Serge Belongie}{sjb344@cornell.edu}{2}{3}
\addauthor{Truong Nguyen}{tqn001@eng.ucsd.edu}{1}

\addinstitution{
 University of California San Diego\\
 La Jolla, CA, USA
}
\addinstitution{
 Cornell University\\
 Ithaca, NY, USA
}
\addinstitution{
 Cornell Tech\\
 New York, NY, USA
}

\runninghead{Tripathi \etal}{Refining Video Object Detection with RNN}

\def\etal{\emph{et al}\bmvaOneDot}

\begin{document}

\maketitle

\begin{abstract}

Given the vast amounts of video available online,
and recent breakthroughs in object detection with static images,
object detection in video offers a promising new frontier.  
However, motion blur and compression artifacts cause substantial frame-level variability,
even in videos that appear smooth to the eye.
Additionally, video datasets tend to have sparsely annotated frames. 
We present a new framework 
for improving object detection in videos
that captures temporal context 
and encourages consistency of predictions. 
First, we train a \emph{pseudo-labeler}, 
that is, a domain-adapted convolutional neural network for object detection.
The pseudo-labeler is first trained individually 
on the subset of labeled frames, 
and then subsequently applied to all frames.
Then we train a recurrent neural network
that takes as input sequences of pseudo-labeled frames
and optimizes an objective 
that encourages both accuracy on the target frame and consistency across consecutive frames.
The approach incorporates
strong supervision of target frames,
weak-supervision on context frames,
and regularization via a smoothness penalty. 
Our approach achieves 
mean Average Precision (mAP) of $68.73$, 
an improvement of $7.1$
over the strongest image-based baselines 
for the Youtube-Video Objects dataset.
Our experiments demonstrate that neighboring frames can provide valuable information,
even absent labels.

\end{abstract}

\section{Introduction}
\label{sec:intro}
Despite the immense popularity and availability of online video content via outlets such as Youtube and Facebook,
most work on object detection  
focuses on static images.
Given the breakthroughs of deep convolutional neural networks 
for detecting objects in static images, 
the application of these methods to video 
might seem straightforward.
However, motion blur and compression artifacts 
cause substantial frame-to-frame variability,
even in videos that appear smooth to the eye.
These attributes complicate prediction tasks 
like classification and localization.
Object-detection models trained on images 
tend not to perform competitively 
on videos owing to domain shift factors \cite{KalogeitonFS15}. 
Moreover, object-level annotations 
in popular video data-sets 
can be extremely sparse, 
impeding the development 
of better video-based object detection models. 

Girshik \etal \cite{RCNN_girshick14CVPR} demonstrate that even given scarce labeled training data, 
high-capacity convolutional neural networks 
can achieve state of the art detection performance 
if first pre-trained on a related task with abundant training data,
such as 1000-way ImageNet classification.
Followed the pretraining, 
the networks can be fine-tuned to a related but distinct domain. 
Also relevant to our work, 
the recently introduced models Faster R-CNN \cite{Faster_RCNN_RenHG015} and You Look Only Once (YOLO) \cite{YOLO_RedmonDGF15}
unify the tasks of classification and localization. These methods, which are accurate and efficient,
propose to solve both tasks through a single model,
bypassing the separate object proposal methods 
used by R-CNN \cite{RCNN_girshick14CVPR}. 

In this paper, we introduce a method
to extend unified object recognition and localization 
to the video domain.
Our approach applies transfer learning 
from the image domain to video frames.
Additionally, we present a novel recurrent neural network (RNN) method
that refines predictions by exploiting contextual information in neighboring frames.
In summary, we contribute the following:

\begin{itemize}
\item A new method for refining a video-based object detection consisting of two parts: (i) a \emph{pseudo-labeler}, which assigns provisional labels 
to all available video frames.
(ii) A recurrent neural network, 
which reads in a sequence of provisionally labeled frames, using the contextual information to output refined predictions.

\item An effective training strategy utilizing (i) category-level weak-supervision at every time-step, (ii) localization-level strong supervision at final time-step (iii) a penalty encouraging prediction smoothness at consecutive time-steps, and (iv) similarity constraints between \emph{pseudo-labels} and prediction output at every time-step.

\item An extensive empirical investigation demonstrating that on the YouTube Objects \cite{youtube-Objects} dataset,
our framework achieves mean average precision (mAP) of $68.73$ on test data, 
compared to a best published result of $37.41$ \cite{Tripathi_WACV16} and $61.66$
for a domain adapted YOLO network \cite{YOLO_RedmonDGF15}.

\end{itemize}
\section{Methods}
\label{sec:method}

In this work, 
we aim to refine object detection in video 
by utilizing contextual information
from neighboring video frames. 
We accomplish this
through a two-stage process.
First, we train a \emph{pseudo-labeler}, 
that is, a domain-adapted convolutional neural network for object detection,
trained individually on the labeled video frames. 
Specifically, we fine-tune the YOLO object detection network \cite{YOLO_RedmonDGF15}, 
which was originally trained for the 20-class PASCAL VOC \cite{PASCAL_VOC} dataset, 
to the Youtube-Video \cite{youtube-Objects} dataset. 

When fine-tuning to the 10 sub-categories 
present in the video dataset, 
our objective is to minimize 
the weighted squared detection loss 
(equation \ref{eqn:obj_det_loss}) 
as specified in YOLO \cite{YOLO_RedmonDGF15}.
While fine-tuning, we learn only the parameters 
of the top-most fully-connected layers,
keeping the $24$ convolutional layers and $4$ max-pooling layers unchanged. 
The training takes roughly 50 epochs to converge, using the RMSProp \cite{RMSProp} optimizer 
with momentum of $0.9$ and a mini-batch size of $128$. 

As with YOLO \cite{YOLO_RedmonDGF15}, 
our fine-tuned $pseudo-labeler$ 
takes $448 \times 448$ frames as input  
and regresses on category types and locations of possible objects
at each one of $S \times S$ non-overlapping grid cells.
For each grid cell, 
the model outputs class conditional probabilities 
as well as $B$ bounding boxes
and their associated confidence scores. 
As in YOLO, we consider a \emph{responsible} bounding box for a grid cell 
to be the one among the $B$ boxes for which the predicted area and the ground truth area 
shares the  maximum Intersection Over Union. 
During training, we simultaneously optimize classification and localization error 
(equation \ref{eqn:obj_det_loss}).
For each grid cell, 
we minimize the localization error 
for the \emph{responsible} bounding box 
with respect to the ground truth 
only when an object appears in that cell.  

Next, we train a Recurrent Neural Network (RNN),
with Gated Recurrent Units (GRUs) \cite{Cho14_GRU}.
This net takes as input 
sequences of \emph{pseudo-labels},  
optimizing an objective 
that encourages both accuracy on the target frame 
and consistency across consecutive frames. 
Given a series of \emph{pseudo-labels} $\mathbf{x}^{(1)}, ..., \mathbf{x}^{(T)}$, 
we train the RNN to generate improved predictions 
$\hat{\mathbf{y}}^{(1)}, ..., \hat{\mathbf{y}}^{(T)}$ 
with respect to the ground truth $\mathbf{y}^{(T)}$ 
available only at the final step in each sequence. 
Here, $t$ indexes sequence steps and $T$ denotes the length of the sequence. 
As output, we use a fully-connected layer  
with a linear activation function, 
as our problem is regression. 
In our final experiments, 
we use a $2$-layer GRU with $150$ nodes per layer, hyper-parameters determined on validation data.

The following equations 
define the forward pass through a GRU layer,
where $\mathbf{h}^{(t)}_l$ denotes the layer's output at the current time step, and $\mathbf{h}^{(t)}_{l-1}$ denotes the previous layer's output at the same sequence step: 
\begin{equation} \label{eqn:GRU}
\begin{aligned}
\mathbf{r}^{(t)}_l &= \sigma(\mathbf{h}^{(t)}_{l-1}W^{xr}_l + \mathbf{h}^{(t-1)}_lW^{hr}_l + \mathbf{b}^r_l)\\
\mathbf{u}^{(t)}_l &= \sigma(\mathbf{h}^{(t)}_{l-1}W^{xu}_l + \mathbf{h}^{(t-1)}_lW^{hu}_l + \mathbf{b}^u_l)\\
\mathbf{c}^{(t)}_l &= \sigma(\mathbf{h}^{(t)}_{l-1}W^{xc}_l + r_t \odot(\mathbf{h}^{(t-1)}_lW^{hc}_l) + \mathbf{b}^c_l)\\
\mathbf{h}^{(t)}_l &= (1-\mathbf{u}^{(t)}_l)\odot \mathbf{h}^{(t-1)}_l + \mathbf{u}^{(t)}_l\odot \mathbf{c}^{(t)}_l
\end{aligned}
\end{equation}
Here, $\sigma$ denotes an element-wise logistic function and $\odot$ is the (element-wise) Hadamard product. 
The reset gate, update gate, and candidate hidden state are denoted by $\textbf{r}$, $\textbf{u}$, and $\textbf{c}$ respectively.
For $S = 7$ and $B=2$, 
the pseudo-labels $\mathbf{x}^{(t)}$ and prediction $\hat{\mathbf{y}}^{(t)}$ both lie in $\mathbb{R}^{1470}$. 
\squeezeup
\subsection{Training}
We design an objective function (Equation \ref{eqn:objective}) that accounts 
for both accuracy at the target frame 
and consistency of predictions across
adjacent time steps in the following ways:  
\begin{equation} \label{eqn:objective}
\mbox{loss} = \mbox{d\_loss} + \alpha \cdot \mbox{s\_loss} + \beta \cdot \mbox{c\_loss} + \gamma \cdot \mbox{pc\_loss}
\end{equation}
Here, d\_loss, s\_loss, c\_loss and pc\_loss stand for detection\_loss, similarity\_loss, category\_loss and prediction\_consistency\_loss described in the following sections. 
The values of the hyper-parameters $\alpha=0.2$, $\beta=0.2$ and $\gamma=0.1$ 
are chosen based on the detection performance on the validation set. 
The training converges in 80 epochs 
for parameter updates using RMSProp \cite{RMSProp} and momentum $0.9$. 
During training we use a mini-batch size of $128$ 
and sequences of length $30$.

\subsubsection{Strong Supervision at Target Frame}
On the final output, 
for which the ground truth classification and localization is available, 
we apply a multi-part object detection loss as described in YOLO \cite{YOLO_RedmonDGF15}.
\squeezeup
\begin{equation} \label{eqn:obj_det_loss}
\begin{aligned}
\mbox{detection\_loss} &= \lambda_{coord}\sum^{S^2}_{i=0}\sum^{B}_{j=0}\mathbbm{1}^{obj}_{ij}\big(\mathit{x}^{(T)}_i - \hat{\mathit{x}}^{(T)}_i\big)^2 + \big(\mathit{y}^{(T)}_i - \hat{\mathit{y}}^{(T)}_i\big)^2 \\
& + \lambda_{coord}\sum^{S^2}_{i=0}\sum^{B}_{j=0}\mathbbm{1}^{obj}_{ij}\big(\sqrt{w_i}^{(T)} - \sqrt{\hat{w}^{(T)}_i}\big)^2 + \big (\sqrt{h_i}^{(T)} - \sqrt{\hat{h}^{(T)}_i} \big)^2 \\
& + \sum^{S^2}_{i=0}\sum^{B}_{j=0}\mathbbm{1}^{obj}_{ij}(\mathit{C}_i - \hat{\mathit{C}_i})^2 \\
& + \lambda_{noobj}\sum^{S^2}_{i=0}\sum^{B}_{j=0}\mathbbm{1}^{noobj}_{ij}\big(\mathit{C}^{(T)}_i - \hat{\mathit{C}}^{(T)}_i\big)^2 \\
& + \sum^{S^2}_{i=0}\mathbbm{1}^{obj}_{i}\sum_{c \in classes}\big(p_i^{(T)}(c) - \hat{p_i}^{(T)}(c)\big)^2
\end{aligned}
\end{equation}
where $\mathbbm{1}^{obj}_{i}$ denotes 
if the object appears in cell $i$ 
and $\mathbbm{1}^{obj}_{ij}$ 
denotes that $j$th bounding box predictor 
in cell $i$ 
is \emph{responsible} for that prediction. 
The loss function penalizes classification 
and localization error differently 
based on presence or absence of an object 
in that grid cell. 
$x_i, y_i, w_i, h_i$ corresponds 
to the ground truth bounding box center coordinates, width and height for objects in grid cell (if it exists) and $\hat{x_i}, \hat{y_i}, \hat{w_i}, \hat{h_i}$ stand for the corresponding predictions. 
$C_i$ and $\hat{C_i}$ denote confidence score of \emph{objectness} at grid cell $i$ for ground truth and prediction.  
$p_i(c)$ and $\hat{p_i}(c)$ 
stand for conditional probability 
for object class $c$ at cell index $i$ 
for ground truth and prediction respectively.  
We use similar settings for YOLO's object detection loss minimization and use values of $\lambda_{coord} = 5$ and $\lambda_{noobj} = 0.5$. 
\squeezeup 

\subsubsection{Similarity between \emph{Pseudo-labels} and Predictions}
Our objective function also includes 
a regularizer that penalizes the dissimilarity between \emph{pseudo-labels} and the prediction at each time frame $t$.
\squeezeup
\begin{equation} \label{auto_enc_loss} 
\mbox{similarity\_loss} = 
\sum^T_{t=0}\sum^{S^2}_{i=0}\hat{C}^{(t)}_i\Big(\mathbf{x}^{(t)}_i - \hat{\mathbf{y}_i}^{(t)}   \Big)^2
\end{equation}
Here, $\mathbf{x}^{(t)}_i$ and $\hat{\mathbf{y}_i}^{(t)}$ denote the \emph{pseudo-labels} and predictions corresponding to the $i$-th grid cell at $t$-th time step respectively. We perform minimization of the square loss weighted by the predicted confidence score at the corresponding cell. 

\subsubsection{Object Category-level Weak-Supervision}
Replication of the static target at each sequential step has been shown to be effective in \cite{LiptonKEW15, yue2015beyond, dai2015semi}.
Of course, with video data, 
different objects may move 
in different directions and speeds. 
Yet, within a short time duration,
we could expect all objects to be present. 
Thus we employ target replication for classification but not localization objectives.

We minimize the square loss 
between the categories 
aggregated over all grid cells 
in the ground truth $\mathbf{y}^{(T)}$ 
at final time step $T$ and predictions $\hat{\mathbf{y}}^{(t)}$ 
at all time steps $t$. 
Aggregated category from the ground truth 
considers only the cell indices 
where an object is present. 
For predictions, contribution of cell $i$ 
is weighted by its predicted confidence score $\hat{C}^{(t)}_i$. 
Note that cell indices with positive detection 
are sparse. 
Thus, we consider the confidence score of each cell while minimizing the aggregated category loss.
\squeezeup
\begin{equation} \label{category_supervision}
\mbox{category\_loss} = 
\sum^T_{t=0}\bigg(\sum_{c \in classes} \Big(\sum^{S^2}_{i=0} \hat{C}^{(t)}_i\big(\hat{p}^{(t)}_i(c)\big) -  \sum^{S^2}_{i=0}\mathbbm{1}^{obj^{(T)}}_i \big(p_i^{(T)}(c)\big)\Big) \bigg)^2
\end{equation}

\subsubsection{Consecutive Prediction Smoothness} 
Additionally, we regularize the model 
by encouraging smoothness of predictions 
across consecutive time-steps.
This makes sense intuitively
because we assume that objects
rarely move rapidly from one frame to another. 
\squeezeup
\begin{equation} \label{prediction_smoothness} 
\mbox{prediction\_consistency\_loss} = 
\sum^{T-1}_{t=0}\Big(\hat{\mathbf{y}_i}^{(t)} - \hat{\mathbf{y}_i}^{(t+1)}   \Big)^2
\end{equation}

\squeezeup
\subsection{Inference}
The recurrent neural network predicts output at every time-step. 
The network predicts $98$ bounding boxes 
per video frame 
and class probabilities 
for each of the $49$ grid cells. 
We note that for every cell, 
the net predicts class conditional probabilities  
for each one of the $C$ categories 
and $B$ bounding boxes. 
Each one of the $B$ predicted bounding boxes 
per cell 
has an associated \emph{objectness} confidence score. 
The predicted confidence score
at that grid is the maximum among the boxes.
The bounding box with the highest score 
becomes the \emph{responsible} prediction 
for that grid cell $i$.      

The product of class conditional probability $\hat{p}^{(t)}_i(c)$ for category type $c$ and \emph{objectness} confidence score $\hat{C}^{(t)}_i$ at grid cell $i$, 
if above a threshold, infers a detection. 
In order for an object of category type $c$ 
to be detected for $i$-th cell 
at time-step $t$, 
both the class conditional probability $\hat{p}^{(t)}_i(c)$ and \emph{objectness score} $\hat{C}^{(t)}_i$ must be reasonably high. 

Additionally, we employ Non-Maximum Suppression (NMS) to winnow multiple high scoring bounding boxes around an object instance 
and produce a single detection for an instance. 
By virtue of YOLO-style prediction, NMS is not critical.

\section{Experimental Results}
\label{sec:results}
In this section, 
we empirically evaluate our model on the popular 
\textbf{Youtube-Objects} dataset,
providing both quantitative results 
(as measured by mean Average Precision) 
and subjective evaluations of the model's performance, considering both successful predictions and failure cases. 

The \textbf{Youtube-Objects} dataset\cite{youtube-Objects} 
is composed of videos collected from Youtube 
by querying for the names of 10 object classes 
of the PASCAL VOC Challenge. 
It contains 155 videos in total 
and between 9 and 24 videos for each class. 
The duration of each video varies 
between 30 seconds and 3 minutes. 
However, only $6087$ frames 
are annotated with $6975$ bounding-box instances. The training and test split is provided.  

\subsection{Experimental Setup}
We implement the domain-adaption of YOLO and 
the proposed RNN model using Theano \cite{Theano2016arXiv160502688short}. 
Our best performing RNN model 
uses two GRU layers of $150$ hidden units each and dropout of probability $0.5$ between layers, 
significantly outperforming domain-adapted YOLO alone. 
While we can only objectively evaluate prediction quality on the labeled frames,
we present subjective evaluations on sequences.

\subsection{Objective Evaluation}
We compare our approach with other methods evaluated on the Youtube-Objects dataset. 
As shown in Table \ref{table:per_category_results} and Table \ref{table:final_mAP},
Deformable Parts Model (DPM) \cite{FelzenszwalbMR_CVPR_2008})-based detector reports \cite{KalogeitonFS15} mean average precision below $30$, 
with especially poor performance 
in some categories such as \emph{cat}.  
The method of Tripathi \etal (VPO) \cite{Tripathi_WACV16} 
uses consistent video object proposals 
followed by a domain-adapted AlexNet classifier (5 convolutional layer, 3 fully connected) \cite{AlexNet12} in an R-CNN \cite{RCNN_girshick14CVPR}-like framework,
achieving mAP of $37.41$. 
We also compare against YOLO ($24$ convolutional layers, 2 fully connected layers),
which unifies the classification and localization tasks,
and achieves mean Average Precision over $55$. 

In our method, we adapt YOLO to generate \emph{pseudo-labels} for all video frames,
feeding them as inputs to the refinement RNN.
We choose YOLO as the \emph{pseudo-labeler}
because it is the most accurate 
among feasibly fast image-level detectors. 
The domain-adaptation improves YOLO's performance, 
achieving mAP of $61.66$. 

Our model with RNN-based prediction refinement, 
achieves superior aggregate mAP to all baselines. 
The RNN refinement model using both input-output similarity, category-level weak-supervision, and prediction smoothness performs best,
achieving $\mbox{68.73}$ mAP.
This amounts to a relative improvement of $\mbox{11.5\%}$ over the best baselines.
Additionally, the RNN improves 
detection accuracy on most individual categories (Table \ref{table:per_category_results}).

\begin{table} \label{table:per_category_results} 
\centering
\footnotesize
\begin{tabular}{lllllllllll}
\multicolumn{11}{c}{\textbf{Average Precision on 10-categories}} \\ \midrule 
    Methods & airplane & bird & boat & car & cat & cow & dog & horse & mbike & train \\ \midrule
    DPM\cite{FelzenszwalbMR_CVPR_2008} & 28.42 & 48.14 & 25.50 & 48.99 & 1.69 & 19.24 & 15.84 & 35.10 & 31.61 & 39.58 \\
    VOP\cite{Tripathi_WACV16} & 29.77 & 28.82 & 35.34 & 41.00 & 33.7 & 57.56 & 34.42 & 54.52 & 29.77 & 29.23 \\
    YOLO\cite{YOLO_RedmonDGF15} & 76.67 & 89.51 & 57.66 & 65.52 & 43.03 & 53.48 & 55.81 & 36.96 & 24.62 & 62.03 \\
    DA YOLO & \textbf{83.89} & \textbf{91.98} & 59.91 & 81.95 & 46.67 & 56.78 & 53.49 & 42.53 & 32.31 & 67.09 \\
 \midrule
    RNN-IOS & 82.78 & 89.51 & 68.02 & \textbf{82.67} & 47.88 & 70.33 & 52.33 & 61.52 & 27.69 & \textbf{67.72} \\
    RNN-WS & 77.78 & 89.51 & \textbf{69.40} & 78.16 & 51.52 & \textbf{78.39} & 47.09 & 81.52 & 36.92 & 62.03 \\
    RNN-PS & 76.11 & 87.65 & 62.16 & 80.69 & \textbf{62.42} & 78.02 & \textbf{58.72} & \textbf{81.77} & \textbf{41.54} & 58.23 \\ \bottomrule
\end{tabular}
\caption{Per-category object detection results   for the Deformable Parts Model (DPM),
Video Object Proposal based AlexNet (VOP), 
image-trained YOLO (YOLO), 
domain-adapted YOLO (DA-YOLO). 
RNN-IOS regularizes on input-output similarity, 
to which RNN-WS adds category-level weak-supervision, 
to which RNN-PS adds a regularizer encouraging prediction smoothness.}
\end{table}

\begin{table}[h] \label{table:final_mAP} 
\centering
\begin{tabular}{llllllll}
\multicolumn{8}{c}{\textbf{mean Average Precision on all categories}} \\ 
\midrule
     Methods & DPM & VOP & YOLO & DA YOLO & RNN-IOS & RNN-WS & RNN-PS\\ \midrule
    mAP & 29.41 & 37.41 & 56.53 & \textbf{61.66} & 65.04 & 67.23 & \textcolor{blue}{\textbf{68.73}}\\ \bottomrule
\end{tabular}
\caption{Overall detection results on Youtube-Objects dataset. Our best model (RNN-PS)  provides $7\%$ improvements over DA-YOLO baseline.}
\end{table}

\squeezeup
\squeezeup

\subsection{Subjective Evaluation}
\begin{figure*}
\begin{center}
	\includegraphics[scale=0.75]{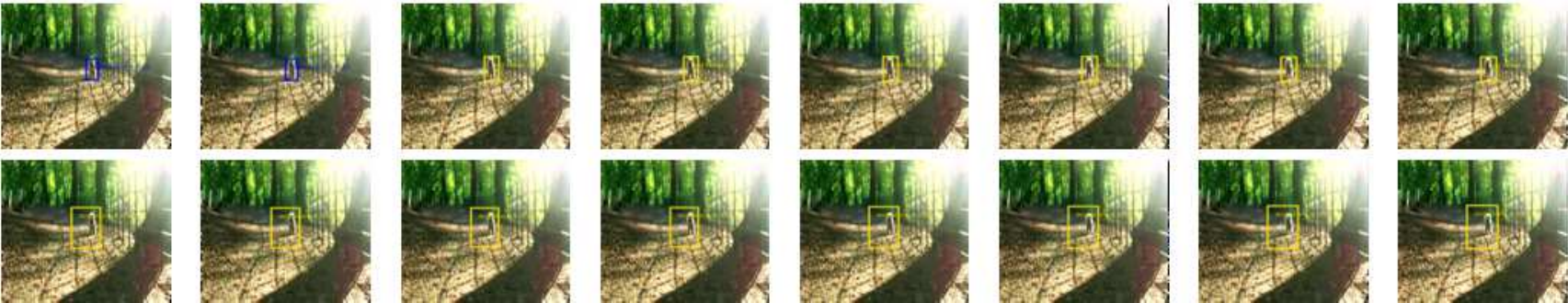}
    \includegraphics[scale=0.75]{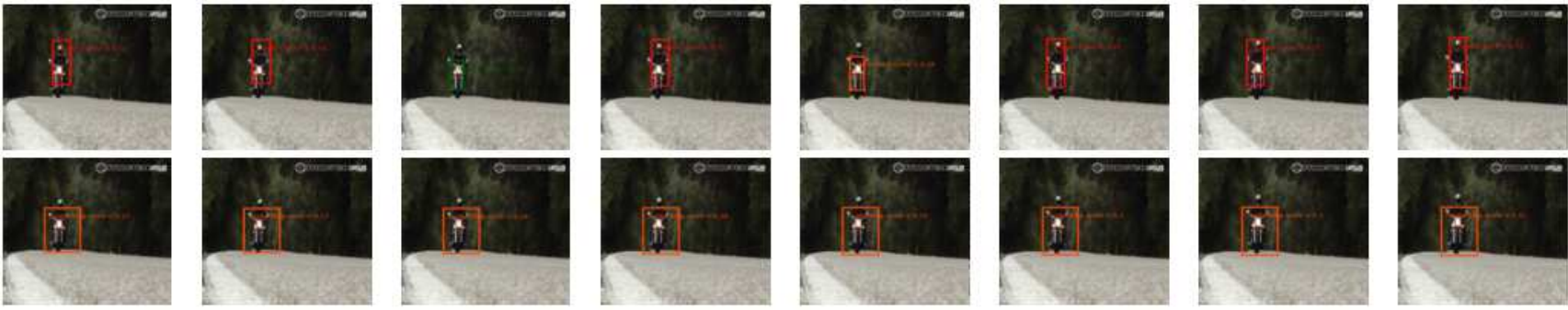}
    \includegraphics[scale=0.75]{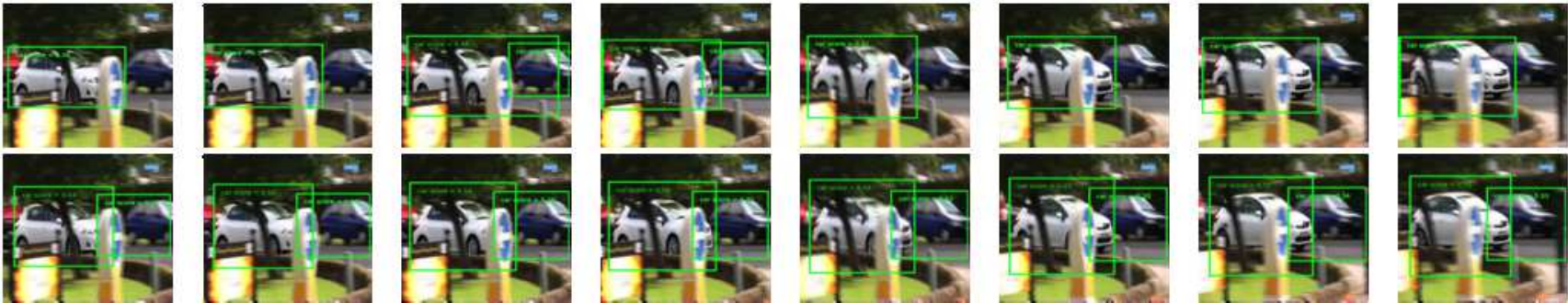}
    \includegraphics[scale=0.75]{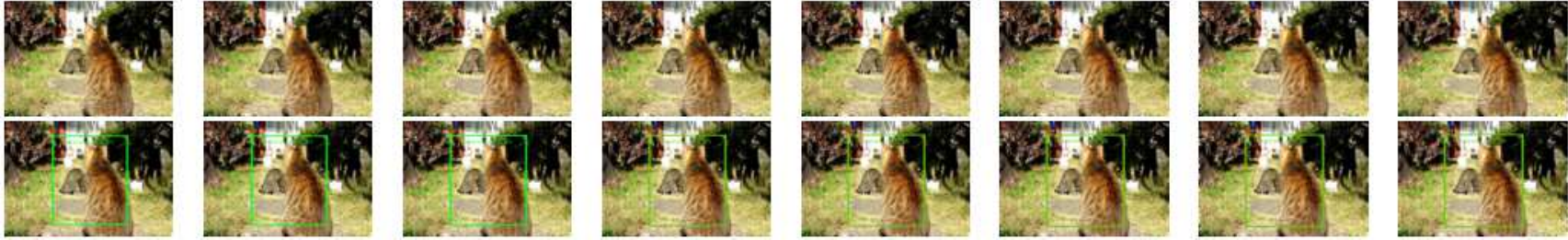}
    \includegraphics[scale=0.75]{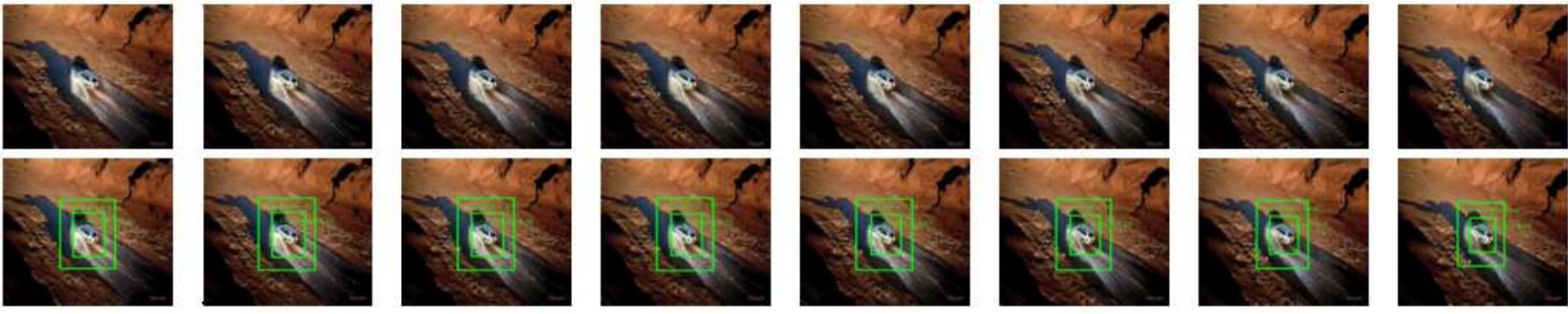}
\end{center}
   \caption{
   Object detection results from the final eight frames of five different test-set sequences. 
   In each pair of rows, 
   the top row shows the \emph{pseudo-labeler} 
   and the bottom row shows the RNN.
   In the first two examples, 
   the RNN consistently predicts correct categories  
   \emph{dog} and \emph{motorbike},
   in contrast to the inconsistent baseline.
   In the third sequence, the RNN 
   correctly predicts multiple instances
   while the \emph{pseudo-labeler} misses one. 
   For the last two sequences, the RNN 
   increases the confidence score, 
   detecting objects missed by the baseline. 
}
\label{fig:subjective1}
\end{figure*}

We provide a subjective evaluation 
of the proposed RNN model in Figure  \ref{fig:subjective1}.
Top and bottom rows in every pair of sequences correspond to \emph{pseudo-labels} and results from our approach respectively. 
While only the last frame 
in each sequence has associated ground truth,
we can observe that the RNN produces more accurate and more consistent predictions across time frames. 
The predictions are consistent 
with respect to classification, 
localization and confidence scores.

In the first example, the RNN 
consistently detects the \emph{dog} 
throughout the sequence, 
even though the \emph{pseudo-labels} 
for the first two frames were wrong (\emph{bird}). In the second example, \emph{pseudo-labels} 
were  \emph{motorbike}, \emph{person}, \emph{bicycle} 
and even \emph{none} at different time-steps. However, our approach consistently predicted \emph{motorbike}. 
The third example shows that the RNN 
consistently predicts both of the cars 
while the \emph{pseudo-labeler}
detects only the smaller car 
in two frames within the sequence. 
The last two examples show how the RNN increases its confidence scores,
bringing out the positive detection 
for \emph{cat} and \emph{car} respectively 
both of which fell below the detection threshold of the \emph{pseudo-labeler}.

\subsection{Areas For Improvement}
The YOLO scheme for unifying classification and localization
\cite{YOLO_RedmonDGF15} 
imposes strong spatial constraints 
on bounding box predictions 
since each grid cell can have only one class. 
This restricts the set of possible predictions,
which may be undesirable 
in the case where many objects 
are in close proximity. 
Additionally, the rigidity of the YOLO model 
may present problems for the refinement RNN,
which encourages smoothness of predictions
across the sequence of frames. 
Consider, for example, an object which moves
slightly but transits from one grid cell to another.
Here smoothness of predictions seems undesirable.

\begin{figure*}
\begin{center}
	\includegraphics[scale=0.75]{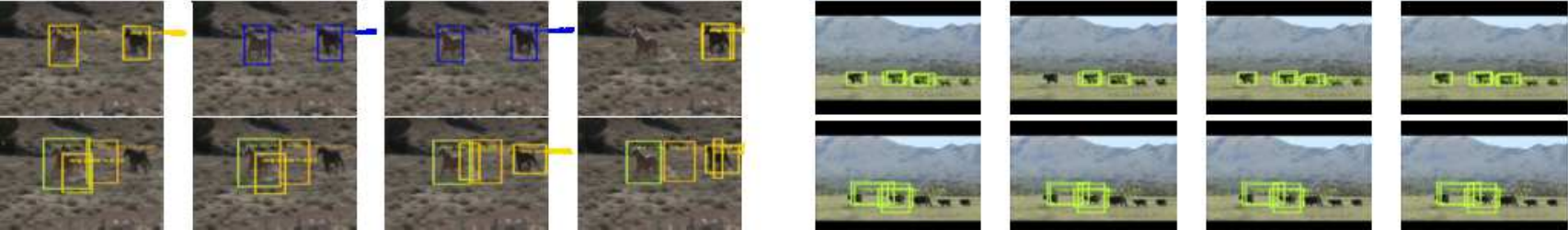}
\end{center}
   \caption{Failure cases for the proposed model. Left: the RNN cannot recover from incorrect \emph{pseudo-labels}. 
Right: RNN localization performs worse than \emph{pseudo-labels} possibly owing to multiple instances of the same object. }
\label{fig:failure_cases}
\squeezeup 
\end{figure*}




Figure \ref{fig:failure_cases} shows some failure cases.  
In the first case, 
the \emph{pseudo-labeler} classifies the instances as \emph{dogs} and even as \emph{birds} in two frames 
whereas the ground truth instances are \emph{horses}.
The RNN cannot recover from the incorrect pseudo-labels.
Strangely, the model increases the confidence score marginally for a different wrong category \emph{cow}. 
In the second case, 
possibly owing to motion and close proximity of multiple instances of the same object category, 
the RNN predicts the correct category but fails 
on localization.
These point to future work to make the framework robust to motion.

The category-level weak supervision in the current scheme
assumes the presence of all objects in nearby frames.
While for short snippets of video 
this assumption generally holds,
it may be violated in case of occlusions, or sudden arrival or departure of objects.
In addition, 
our assumptions regarding the desirability of  
prediction smoothness
can be violated in the case of rapidly moving objects. 



\squeezeup
\section{Related Work}
\label{sec:prior-art}
Our work builds upon a rich literature in both image-level object detection,video analysis, and recurrent neural networks.
Several papers propose ways of using deep convolutional networks for detecting objects \cite{RCNN_girshick14CVPR,fast_RCNN_15,Faster_RCNN_RenHG015, YOLO_RedmonDGF15, SzegedyREA14, Inside_Outside_Net_BellZBG15, DeepID-Net_2015_CVPR, Overfeat_SermanetEZMFL13, CRAFTCVPR16, Gidaris_2015_ICCV}. 
Some approaches classify the proposal regions \cite{RCNN_girshick14CVPR,fast_RCNN_15} into object categories and some other recent methods \cite{Faster_RCNN_RenHG015, YOLO_RedmonDGF15} 
unify the localization and classification stages.   
Kalogeiton \etal \cite{KalogeitonFS15} identifies domain shift factors between still images and videos, necessitating
video-specific object detectors.
To deal with shift factors and sparse object-level annotations in video, 
researchers have proposed several strategies. 
Recently, \cite{Tripathi_WACV16} proposed 
both transfer learning from the image domain to video frames and optimizing for temporally consistent object proposals. 
Their approach is capable of detecting 
both moving and static objects.
However, the object proposal generation step 
that precedes classification is slow. 

Prest \etal 
\cite{Weak_obj_from_videoPrestLCSF12},
utilize weak supervision for object detection 
in videos
via category-level annotations of frames,
absent localization ground truth.
This method assumes that the target object is moving, outputting a spatio-temporal tube that captures this most salient moving object.
This paper, however, does not consider
context within video for detecting multiple objects.

A few recent papers \cite{DeepID-Net_2015_CVPR, Inside_Outside_Net_BellZBG15} identify the important role of context in visual recognition. 
For object detection in images, Bell \etal \cite{Inside_Outside_Net_BellZBG15} 
use spatial RNNs 
to harness contextual information, 
showing large improvements on PASCAL VOC \cite{PASCAL_VOC} and Microsoft COCO \cite{COCOLinMBHPRDZ14}
object detection datasets.
Their approach adopts proposal generation followed by classification framework. 
This paper exploits spatial, 
but not temporal context. 

Recently, Kang \etal \cite{KangCVPR16} introduced tubelets with convolutional neural networks (T-CNN) for detecting objects in video. 
T-CNN uses spatio-temporal tubelet proposal generation 
followed by the classification and re-scoring, 
incorporating temporal and contextual information from tubelets obtained in videos. 
T-CNN won the recently introduced ImageNet 
object-detection-from-video (VID) task with provided densely annotated video clips.
Although the method is effective for densely annotated training data, 
it's behavior for sparsely labeled data is not evaluated. 


By modeling video as a time series,
especially via GRU \cite{Cho14_GRU} or LSTM RNNs\cite{LSTM_Hochreiter_97},
several papers 
demonstrate improvement
on visual tasks including video classification \cite{yue2015beyond},
activity recognition \cite{LongTermRecurrentDonahueHGRVSD14}, 
and human dynamics \cite{Fragkiadaki_2015_ICCV}. 
These models generally aggregate CNN features 
over tens of seconds, which forms the input to an RNN. 
They perform well for global description tasks
such as classification \cite{yue2015beyond,LongTermRecurrentDonahueHGRVSD14} but require large annotated datasets.
Yet, detecting multiple generic objects 
by explicitly modeling video 
as an ordered sequence 
remains less explored. 

Our work differs from the prior art in a few distinct ways. 
First, this work is the first, to our knowledge, to demonstrate the capacity of
RNNs to improve localized object detection in videos.
The approach may also be the first 
to refine the object predictions of frame-level models.
Notably, our model produces significant improvements even on a small dataset with sparse annotations.

\squeezeup
\squeezeup
\section{Conclusion}
We introduce a framework 
for refining object detection in video. 
Our approach extracts contextual information from neighboring frames,
generating predictions 
with state of the art accuracy
that are also temporally consistent. 
Importantly, our model 
benefits from context frames
even when they lack ground truth annotations. 

For the recurrent model,
we demonstrate an efficient 
and effective training strategy 
that simultaneously employs 
localization-level strong supervision, 
category-level weak-supervision, 
and a penalty 
encouraging smoothness of predictions 
across adjacent frames. 
On a video dataset with sparse object-level annotation, 
our framework proves effective,
as validated by extensive experiments. 
A subjective analysis of failure cases 
suggests that the current approach 
may struggle most on cases 
when multiple rapidly moving objects 
are in close proximity. 
Likely, the sequential smoothness penalty is not optimal for such complex dynamics. 

Our results point to several promising directions for future work. First, recent state of the art results for video classification show that longer sequences help in global inference.
However, the use of longer sequences for localization remains unexplored. 
We also plan to explore methods 
to better model local motion information 
with the goal of improving localization of
multiple objects in close proximity. 
Another promising direction, we would like
to experiment with loss functions 
to incorporate specialized handling
of classification and localization objectives.


\bibliography{bmvc_review}
\end{document}